\documentclass[10pt,twocolumn,letterpaper]{article}

\usepackage{wacv}
\usepackage{times}
\usepackage{epsfig}
\usepackage{graphicx}
\usepackage{amsmath}
\usepackage{amssymb}
\usepackage{adjustbox}
\usepackage{rotating}
\usepackage{multirow}
\usepackage{mathtools}

\usepackage{caption}
\usepackage{subcaption}
\usepackage{algorithmic}
\usepackage[ruled,vlined]{algorithm2e}
\usepackage{footnote}

\DeclarePairedDelimiter\floor{\lfloor}{\rfloor}

\graphicspath{ {figures/} }

\def\wacvPaperID{1329} % Enter the WACV Paper ID here

\wacvfinalcopy % *** Uncomment this line for the final submission

\ifwacvfinal
\def\assignedStartPage{9876} % *** Enter the assigned starting page number (instead of 9876)
\fi

\ifwacvfinal
\usepackage[breaklinks=true,bookmarks=false]{hyperref}
\else
\usepackage[pagebackref=true,breaklinks=true,colorlinks,bookmarks=false]{hyperref}
\fi

\ifwacvfinal
\else
\fi

\begin{document}

%%%%%%%%% TITLE
\title{Multi Projection Fusion for Real-time Semantic Segmentation of 3D LiDAR Point Clouds}

\author{Yara Ali Alnaggar\thanks{equal contribution} \hspace{1cm} Mohamed Afifi\footnotemark[1] \hspace{1cm} Karim Amer \hspace{1cm} Mohamed ElHelw\\
Center for Informatics Science, Nile University\\
Giza, Egypt\\
{\tt\small \{y.ali, moh.afifi, k.amer, melhelw\}@nu.edu.eg}
%\and
}

\maketitle
%\thispagestyle{empty}

%%%%%%%%% ABSTRACT
\begin{abstract}
Semantic segmentation of 3D point cloud data is essential for enhanced high-level perception in autonomous platforms. Furthermore, given the increasing deployment of LiDAR sensors onboard of cars and drones, a special emphasis is also placed on non-computationally intensive algorithms that operate on mobile GPUs. Previous efficient state-of-the-art methods relied on 2D spherical projection of point clouds as input for 2D fully convolutional neural networks to balance the accuracy-speed trade-off. This paper introduces a novel approach for 3D point cloud semantic segmentation that exploits multiple projections of the point cloud to mitigate the loss of information inherent in single projection methods. Our Multi-Projection Fusion (MPF) framework analyzes spherical and bird's-eye view projections using two separate highly-efficient 2D fully convolutional models then combines the segmentation results of both views. The proposed framework is validated on the SemanticKITTI dataset where it achieved a mIoU of 55.5 which is higher than state-of-the-art projection-based methods RangeNet++ \cite{milioto2019rangenet++} and PolarNet \cite{zhang2020polarnet} while being 1.6x faster than the former and 3.1x faster than the latter.
\end{abstract}

% \begin{figure}[t]
%      \centering
%          \includegraphics[width=0.85\linewidth]{qualitaitve/rect_label_rect.jpg}
%          \includegraphics[width=0.85\linewidth]{qualitaitve/rect_ours_rect.jpg}
%          \includegraphics[width=0.85\linewidth]{qualitaitve/rect_rangenet_rect.jpg}

%         \caption{Our MPF framework outperforms RangeNet when segmenting objects that are far from LiDAR. This labeled scan belongs to SemanticKITTI validation sequence (8), where each color represents a class. Top: Scan ground truth where the object marked by a dashed rectangle belongs to ``other-vehicle" class. Middle: Our framework classifies correctly that distant object as ``other-vehicle". Bottom: RangeNet labels the same object falsely as ``car'' class.}
%         \label{fig:qualitative_result}
% \end{figure}
\begin{figure}[t]
     \centering
         \includegraphics[width=0.95\linewidth]{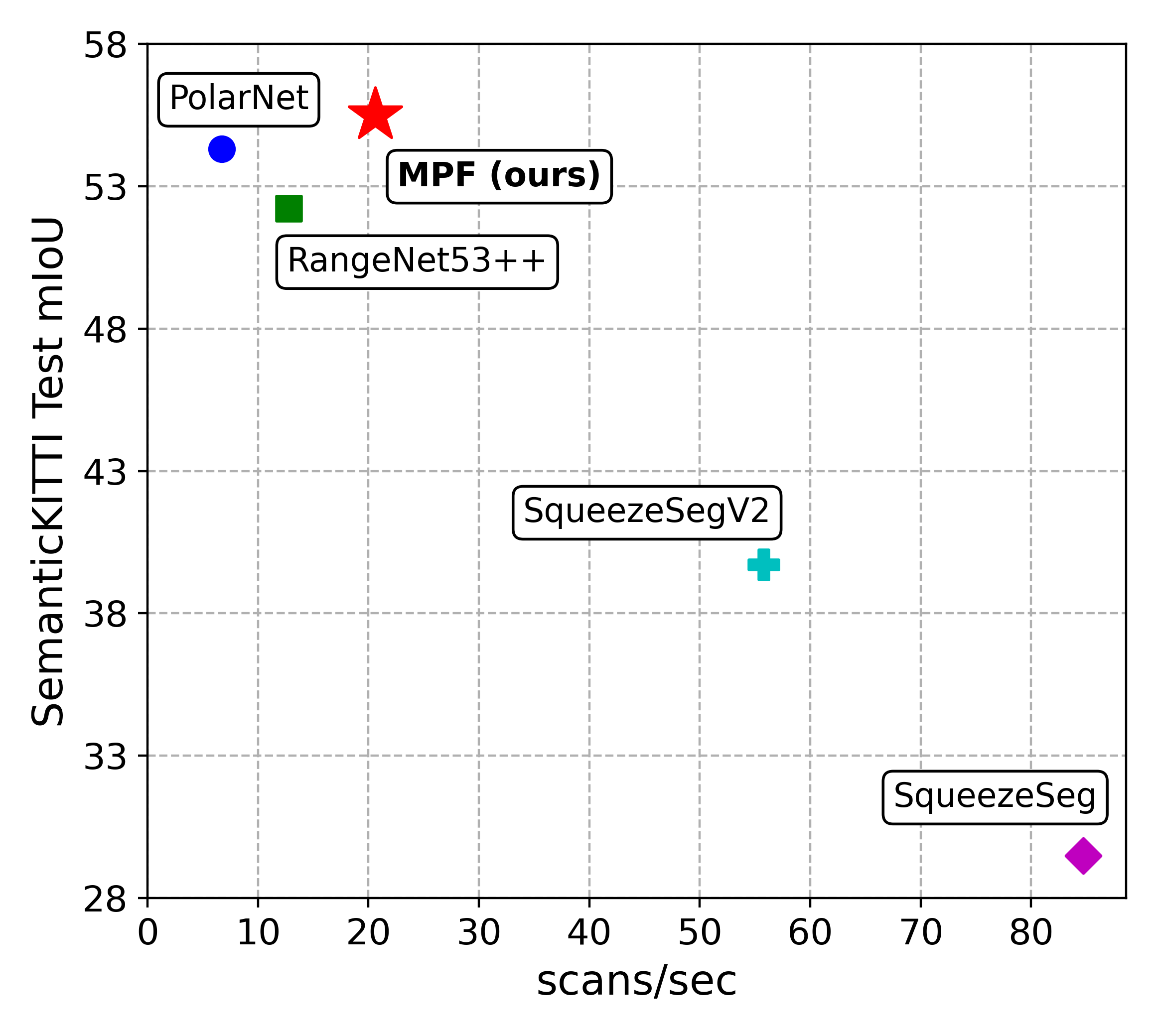}
        \caption{Computed scans per second vs. mIoU score on the test set of SemanticKITTI dataset using state-of-the-art projection-based methods. Our framework achieves the highest mIoU score; in addition, it is \textbf{3.1} and \textbf{1.6} times faster than PolarNet and RangeNet53++, respectively.}
        \label{fig:qualitative_result}
\end{figure}

%%%%%%%%% BODY TEXT
\section{Introduction}

Currently, Light Detection and Ranging (LiDAR) sensors are widely used in autonomous navigation systems where captured 3D point cloud data provides a rich source of information on the surrounding scene. Analyzing such data with deep learning models has gained a lot of attention in the research community especially for extracting semantic information to improve navigation accuracy and safety. In this case, semantic segmentation algorithms assign a label for each point in the 3D point cloud representing different classes of objects in the scene.

Convolutional Neural Networks (CNNs) have achieved state-of-the-art results in semantic segmentation tasks with fully convolutional architectures trained on huge amounts of labelled RGB data and making clever use of transfer learning. However, the same success has not yet been achieved in semantic segmentation of point cloud data due to lack of large annotated datasets. Furthermore, since point cloud semantic segmentation models are typically deployed on devices with limited computational capabilities onboard of mobile platforms (e.g. cars or drones), there is a need for high throughput while sustaining high accuracy to ensure the platform has enough time to make correct decisions.

There are two main approaches in the literature to tackle the task of semantic segmentation of 3D point clouds. The first applies 3D CNN models either on the raw cloud data points \cite{qi2017pointnet} or after transforming the points into 3D volumetric grid representations \cite{tchapmi2017segcloud}. This incurs high computational costs \cite{behley2019iccv} and hence not suitable for real-time systems. The second approach applies 2D CNN models to 2D projections of the 3D point cloud based on either bird's-eye view \cite{radi2019volmap} or spherical view \cite{8462926}. Currently, state-of-the-art methods such as RangeNet++ \cite{milioto2019rangenet++} applies a Fully Convolutional Neural Network (FCNN) on a spherical projection of the point cloud. However, there is an inevitable loss of information due to the projection operation which can limit model performance especially for distant points. In this paper, we introduce a novel framework for enhanced online semantic segmentation of 3D point clouds by incorporating multi-view projections of the same point cloud which results in an improved performance compared to single-projection models while attaining real-time performance. 
The contributions of this work can be summarized as follows. First, a novel MPF framework that utilizes multi-view projections and fusion of input point cloud to make up for the loss of information inherent in single projection methods. Second, the MPF framework processes spherical and bird's-eye projections using two independent models that can be selected to achieve optimum performance for a given platform (road vehicle, aerial drone, etc.) and/or deployed on separate GPUs. Third, the framework is scalable and, despite using only two projections in the current work, it can be directly extended to exploit multiple projections.

Incorporating information from multiple projections of 3D data has been used before in other domains to improve performance. Mortazi et al. \cite{mortazi2017cardiacnet} used a single 2D encoder-decoder CNN for CT-scan segmentation to parse all 2D slices in X, Y and Z directions. Chen et al. \cite{chen2017multi} used multiple 2D encoder CNN on spherical and bird's-eye views for the task of 3D object detection in point cloud data. However, to the best of our knowledge this setup has not been used before in semantic segmentation of 3D point clouds. This is primarily due to the added computational overhead of having the same complex network architecture for multiple views and back projection of results to the original point cloud space to compute point-level predictions (unlike \cite{chen2017multi} who only outputs 3D bounding boxes). The paper is organized as follows: Section 2 overviews related work, Section 3 describes the proposed framework and Section 4 presents obtained experimental results as well as an ablation study of our framework. Section 5 concludes the paper and points out future work directions.

\begin{figure*}
\begin{center}
% \fbox{\rule{0pt}{2in} \rule{.9\linewidth}{0pt}}
\includegraphics[width=\linewidth]{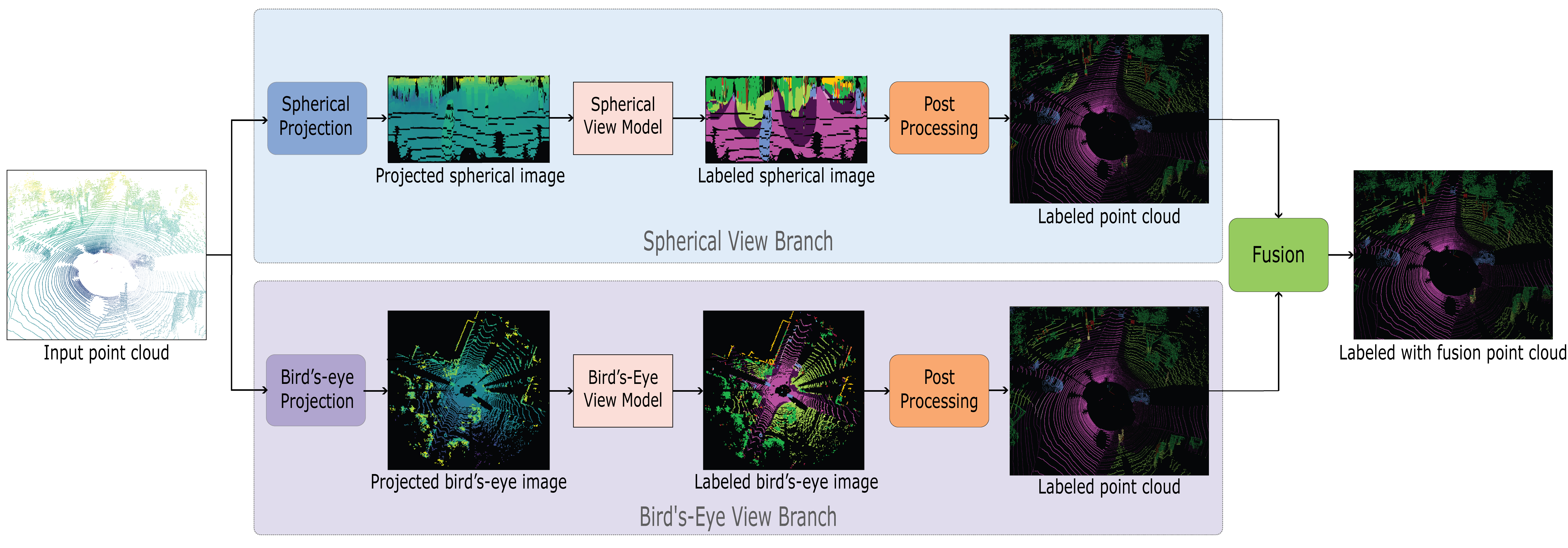}
\end{center}
   \caption{Proposed MPF framework overview. The framework takes as input a 3D point cloud that undergoes a series of operations in the Spherical View Branch and the Bird'e-Eye View branch. Each branch is composed of three main processing blocks where the first block transforms the 3D point point cloud into its respective 2D projection. The second block, Spherical or Bird's-Eye View Model, predicts segmentation of the projected 2D image with a FCNN model where each view has its own model. The third block, Post Processing, further processes the semantically-segmented projected view and assigns to each point in the input cloud its corresponding softmax probabilities. Finally, information from the two branches are fused by the Fusion block to produce the final semantic label of each point in the 3D point cloud.}
\label{fig:pipepline}
\end{figure*}

%------------------------------------------------------------------------
\section{Related Work}

\subsection{Semantic Image Segmentation}

Semantic image segmentation has attracted a lot of attention in recent years following the success of deep CNN models, such as AlexNet \cite{krizhevsky2012imagenet}, VGGNet \cite{simonyan2014very}, and ResNet \cite{he2016deep} in image classification. The task aims at predicting pixel-level classification labels in order to have more precise information on objects in input images.  One of the earliest work in this area is based on using Fully Convolutional Neural Networks (FCNNs) \cite{long2015fully} where the model can assign a label for each pixel in the image in a single forward pass by extracting features using multi-layer encoder (in this case it was VGG model \cite{simonyan2014very}) and apply up-sampling on these features combined with 1x1 convolution layer to classify each pixel. The idea was extended by Noh et al. \cite{noh2015learning} and Badrinarayanan et al. \cite{badrinarayanan2017segnet} by using a multi-layer decoder to transform extracted features into image space with the needed pixel labels. Ronneberger et al. \cite{ronneberger2015u} introduced the UNet architecture where skip connections between encoder and decoder layers further improve segmentation results. In addition to advances in model architecture design, having large annotated datasets for semantic segmentation tasks, such as Microsoft's COCO \cite{lin2014microsoft}, and utilizing transfer learning from image classification models pre-trained on large datasets such as ImageNet \cite{deng2009imagenet} can significantly improve semantic segmentation accuracy.

\subsection{Semantic Segmentation of 3D Point Cloud}

Most 3D point cloud semantic segmentation algorithms employ a Fully Convolutional Neural Network (FCNN) in a way similar to 2D semantic image segmentation but with the difference of how the FCNN is applied to 3D structures. These algorithms can essentially be grouped into two categories. The first category includes models that use 3D convolutions as in SegCloud \cite{tchapmi2017segcloud} where a 3D FCNN is applied to point cloud after voxelization and transformation into homogeneous 3D grid. However, 3D convolutions are computationally intensive and 3D volumes of point clouds are typically sparse. Other models have special convolution layers for 3d points in order to process a point cloud in its raw format with examples including PointNet++ \cite{qi2017pointnet++}, TangentConv \cite{Tat2018} and KPConv \cite{thomas2019kpconv}. Recent work in RandLA \cite{hu2019randla} improves run-time while maintaining a high segmentation accuracy compared to previously mentioned methods.

The second category of algorithms apply 2D FCNN models after projecting 3D point clouds onto 2D space. In SqueezeSeg \cite{8462926} and SqueezeSegV2 \cite{8793495}, spherical projection is performed on point cloud then 2D encoder-decoder architecture is applied. RangeNet++ \cite{milioto2019rangenet++} improves segmentation results by using a deeper FCNN model and employing post-processing using K-Nearest Neighbour (KNN). Algorithms in this category not only improve inference time but also enhance segmentation accuracy by capitalizing on the success of CNNs in 2D image segmentation. Different projections can also be used as in VolMap \cite{radi2019volmap} which uses Cartesian bird's-eye projection and PolarNet \cite{zhang2020polarnet} which uses polar bird's-eye projection combined with ring convolutions. The proposed framework advances current state-of-the-art of projection based methods in point cloud segmentation by making use of both spherical and Cartesian bird's-eye projections to reduce the loss of information stemming from using a single projection.
Finally, it is worth mentioning that one of the key challenges in 3D point cloud semantic segmentation is the lack of large labeled point cloud datasets. The current benchmark dataset is SemanticKITTI \cite{behley2019iccv} which has around 43K frames collected using 360 Velodyne LiDAR sensor \cite{Velodyne}. Other datasets are either generated synthetically using simulation environments such as Virtual Kitti \cite{gaidon2016virtual} or have small number of samples such as Paris-Lille-3D \cite{roynard2018paris}.

\subsection{Efficient Deep Learning Architectures}

With the impressive results achieved by deep learning models in detection and classification tasks, there is a growing need for efficient models to be deployed on embedded devices and mobile platforms. A number of network architectures that achieve high classification accuracy while having real-time inference have been recently proposed. For instance, MobileNetV1 \cite{howard2017mobilenets} uses depth-wise separable convolutions while ShuffleNet \cite{zhang2018shufflenet} utilizes group convolutions and channel shuffling to reduce computations. MobileNetV2 \cite{sandler2018mobilenetv2} achieves improved  accuracy while maintaining fast inference by using inverted residual blocks. Although these models are designed for classification tasks, they can be used in the context of semantic segmentation as encoders in FCNN models in order to benefit from their efficient architecture.

\subsection{Segmentation Loss}

Segmentation models, regardless of their input, are initially trained using classification losses such as Cross Entropy loss or Focal loss \cite{lin2017focal} because the end goal is to assign a label for each pixel (or point). However, such losses lack the global information of predicted and target object masks. Therefore, research work has been conducted to develop a loss function that penalizes the difference between predicted and ground-truth masks as a whole. Milletari et al. \cite{milletari2016v} developed a soft version of Dice coefficient with continuous probabilities instead of discrete 0 or 1 values while Berman et al. \cite{berman2018lovasz} introduced Lovasz Softmax loss which is a function surrogate approximation of the Jaccard coefficient \cite{jaccard1901etude}. Currently, Lovasz Softmax loss is the state-of-the-art segmentation loss and is usually combined with classification loss to have a penalty on both local and global information.

%-----------------------------------------------------------------------

\section{Proposed Method}
We developed a Multi-Projection Fusion (MPF) framework for semantic segmentation of 3D point clouds that relies on using two 2D projections of the point cloud where each projection is processed by an independent FCNN model. As illustrated in Figure \ref{fig:pipepline}, the proposed pipeline starts by feeding the input point cloud to two branches, one responsible for spherical view projection and the other for bird's-eye view projection. Each branch applies semantic segmentation on the projected point cloud. Subsequently, predictions from the two branches are fused to produce the final prediction. It is assumed that the input point cloud is collected by a LiDAR sensor that returns point coordinates $x,y,z$ values and remission of returned signals $rem$, e.g  Velodyne HDL-64E \cite{Velodyne}. In the following sub-sections, we present details of each block in the two branches of the pipeline.

\begin{table}
\begin{center}
\begin{tabular}{c|c|c|c|c|c} 
\hline
Input       & Operator   & t & c   & n & s  \\ 
\hline
$5\times 64\times w$    & conv2d     & - & 32  & - & 1  \\ 
\hline
$32\times 64\times w$     & bottleneck & 1 & 16  & 1 & 2  \\ 
\hline
$16\times 64\times \frac{w}{2}$   & bottleneck & 6 & 24  & 2 & 2  \\ 
\hline
$24\times 64\times \frac{w}{4}$   & bottleneck & 6 & 32  & 3 & 2  \\ 
\hline
$32\times 64\times \frac{w}{8}$   & bottleneck & 6 & 64  & 4 & 2  \\ 
\hline
$64\times 64\times \frac{w}{16}$  & bottleneck & 6 & 96  & 3 & 1  \\ 
\hline
$96\times 64\times \frac{w}{16}$  & bottleneck & 6 & 160 & 3 & 2  \\ 
\hline
$160\times 64\times \frac{w}{32}$ & bottleneck & 6 & 320 & 1 & 1  \\ 
\hline
$320\times 64\times \frac{w}{4}$  & deconv2d   & - & 96  & - & 8  \\ 
\hline
$96\times 64\times w$     & deconv2d   & - & 32  & - & 4  \\ 
\hline
$32\times 64\times w$     & conv2d     & - & 20  & - & 1  \\
\hline
\end{tabular}
\end{center}
\caption{Spherical-View Model Architecture. The model is based on MobileNetV2 \cite{sandler2018mobilenetv2} with additional 2 deconvolutional layers and one 1x1 convolutional layer as a decoder. $n$ is the repetition number for a sequence of layers in block, $t$ is block expansion factor, $c$ is the number of output channels, $s$ is block stride.}
\label{tab:spher_network}
\end{table}

\subsection{Spherical View Projection}

    This section explains the process of transforming a $360^{\circ}$ point cloud into a 2D spherical image that is fed into subsequent Spherical View Model block, as proposed by \cite{8462926}. At the start, the 3D point cloud is mapped from Euclidean space $(x,y,z)$ to Spherical space $(\theta,\phi,r)$ by applying Equation 1.

\begin{equation}\renewcommand*{\arraystretch}{1.5}
\begin{pmatrix}\theta\\\phi\\r\\\end{pmatrix}
= \begin{pmatrix}arcsin(\frac{z}{\sqrt{x^2+y^2+z^2}})\\arctan(y,x)\\\sqrt{x^2+y^2+z^2}\\\end{pmatrix}
\end{equation}
Subsequently, the points are embedded into a 2D spherical image with dimensions  $(H,W)$ by discretizing points' $\theta$ and $\phi$ angles using Equation 2:
\begin{equation}\renewcommand*{\arraystretch}{1.5}
\begin{pmatrix}u\\v\\\end{pmatrix}
= \begin{pmatrix}\frac{1}{2}[1-\phi\pi^{-1}]w\\
[1-(\theta+f_{up})f^{-1}]h\\\end{pmatrix},
\end{equation}
where $u$ and $v$ represent point indices in the spherical image and $f = f_{up}+ f_{down}$ is the sensor's vertical field-of-view.
The mapping and discretization steps may result in some 3D points sharing the same $u$ and $v$ values. To mitigate this condition, 3D points that are closer to LiDAR are given priority to be represented in the 2D image by ordering the points descendingly based on their range value.  The ordered list of points will be embedded into the 2D spherical image using its corresponding $u$ and $v$ coordinates. 
By the end of this process, the resulting 2D spherical image will have five channels corresponding to distinct point features: $x,y,z,r$ and remission $rem$ which is analogous to RGB images that have three channels, one for each color. 

\subsection{Spherical-View Model}
 The Spherical-View Model is a deep learning segmentation model based on FCNN architecture with encoder and decoder parts. The network encoder utilises MobileNetV2 \cite{sandler2018mobilenetv2} as lightweight backbone that provides real-time performance on mobile devices. The backbone is composed of a sequence of basic building blocks called inverted residual blocks that form bottlenecks with residuals. The first and last bottleneck layer expands and compresses input and output tensors, respectively. The intermediate layers are high-capacity layers responsible for extracting high-level information from the expanded tensors. 
 
 For network decoder, we apply two learnable upsampling layers known as transposed convolution layers. The first layer upsamples the input tensor 8 times and the second layer 4 times. At the end, we add convolution layer and softmax logits to output semantically segmented image. Furthermore, dropout layers are added as regularization. Table \ref{tab:spher_network} provides details of the Spherical-View Model layers.

\subsection{Bird's-Eye View Projection}

\begin{table}
\begin{center}
\begin{tabular}{c|c|c|c} 
\hline
Input       & Operator  & c & s  \\ 
\hline
$4\times256\times256$   & ConvBlock & 64         & 1           \\ 
\hline
$64\times256\times256$ & DownBlock & 128        & 2           \\ 
\hline
$128\times128\times128$ & DownBlock & 256        & 2           \\ 
\hline
$256\times64\times64$   & UpBlock   & 128        & 2           \\ 
\hline
$128\times128\times128$ & UpBlock   & 64         & 2           \\ 
\hline
$64\times256\times256$  & conv2d    & 20         & 1           \\
\hline
\end{tabular}
\end{center}
\caption{Bird's-Eye View Model Architecture. $c$ is the number of output channels and $s$ is layer stride.}
\label{tab:birdeye_network}
\end{table}

The second projection in our framework is the 2D bird's-eye view projection. It uses the $x$ and $y$ coordinates of each point and collapses the 3D cloud along the $z$ dimension. The 3D point cloud is thus projected  on the $x-y$ plane that is discretized using a rectangular grid with a defined width and height. For each cell in the grid, we keep at most one projected point corresponding to the point that has the maximum $z$ value among all points projected onto that cell. Points that get projected outside the boundaries of the grid are discarded. Finally, the grid is converted into a 4-channel image where each pixel in the image represents a cell in the grid. Four cell attributes are extracted to form the 4 channels of the image, namely $x$, $y$, $z$, and remission $rem$.

\subsection{Bird's-Eye View Model}

Although MobileNetV2 \cite{sandler2018mobilenetv2} is highly efficient, using it twice for both views will decrease the overall throughput. Since the MPF framework allows using independent network in each processing branch, we decided to use a network with fewer parameters compared to MobileNetV2 \cite{sandler2018mobilenetv2}. Specifically, a light weight modified version of the UNet \cite{ronneberger2015u} encoder-decoder architecture is used for segmentation of bird's-eye view images. As shown in table \ref{tab:birdeye_network}, the encoder consists of 2 downsampling convolutional blocks and the decoder consists of 2 upsampling convolutional blocks with skip connections between corresponding encoder and decoder blocks. In our experiments it is shown that it is sufficient to use only two blocks in both encoder and decoder which significantly improves network inference efficiency. Each block consists of two 2D convolution layers with kernel size 3 followed by max pooling for encoder and preceded by bi-linear upsampling for decoder. We use 2D batch normalization \cite{ioffe2015batch} followed by ELU \cite{clevert2015fast} non-linearity between successive convolutional layers.

\begin{table*}[t]
\begin{center}
%\adjustbox{\linewidth}{!}
\begin{adjustbox}{width=\linewidth}

\begin{tabular}{l|c|c|c|c|c|c|c|c|c|c|c|c|c|c|c|c|c|c|c|c|c|c|c|c} 
\hline
Model        & \begin{sideways}Input Size\end{sideways} & \begin{sideways}scans/sec\end{sideways} & \multicolumn{1}{l|}{\begin{sideways}\#Params\end{sideways}} & \multicolumn{1}{l|}{\begin{sideways}\#MACs\end{sideways}} & \begin{sideways}mIoU\end{sideways} & \begin{sideways}car\end{sideways} & \begin{sideways}bicycle\end{sideways} & \begin{sideways}motorcycle\end{sideways} & \begin{sideways}truck\end{sideways} & \begin{sideways}other-vehicle\end{sideways} & \begin{sideways}person\end{sideways} & \begin{sideways}bicyclist\end{sideways} & \begin{sideways}motorcyclist\end{sideways} & \begin{sideways}road\end{sideways} & \begin{sideways}parking\end{sideways} & \begin{sideways}sidewalk\end{sideways} & \begin{sideways}other-ground\end{sideways} & \begin{sideways}building\end{sideways} & \begin{sideways}fence\end{sideways} & \begin{sideways}vegetation\end{sideways} & \begin{sideways}trunk\end{sideways} & \begin{sideways}terrain\end{sideways} & \begin{sideways}pole\end{sideways} & \begin{sideways}traffic-sign\end{sideways}  \\ 
\hline
PointNet \cite{qi2017pointnet}  & \multirow{7}{*}{50000pts}                & \multirow{7}{*}{-}                      & \multirow{7}{*}{-}                                          & \multirow{7}{*}{-}                                        &   14.6                               & 46.3                              & 1.3                                   & 0.3                                      & 0.1                                 & 0.8                                         & 0.2                                  & 0.2                                     & 0.0                                        & 61.6                               & 15.8                                  & 35.7                                   & 1.4                                        & 41.4                                   & 12.9                                & 31.0                                     & 4.6                                 & 17.6                                  &2.4                                 & 3.7                                         \\                     
PointNet++ \cite{qi2017pointnet++} &                                       &                                         &                                                             &                                                           &   20.1                               & 53.7                              & 1.9                                   & 0.2                                      & 0.9                                 & 0.2                                         & 0.9                                  & 1.0                                     & 0.0                                        & 72.0                               & 18.7                                  & 41.8                                   & 5.6                                        & 62.3                                   & 16.9                                & 46.5                                     & 13.8                                & 30.0                                  & 6.0                                & 8.9                                         \\
SPGraph \cite{landrieu2018large}   &                                       &                                         &                                                             &                                                           &   20.0                               & 68.3                              & 0.9                                   & 4.5                                      & 0.9                                 & 0.8                                         & 1.0                                  & 6.0                                     & 0.0                                        & 49.5                               & 1.7                                   & 24.2                                   & 0.3                                        & 68.2                                   & 22.5                                & 59.2                                     & 27.2                                & 17.0                                  & 18.3                               & 10.5                                        \\
SPLATNet \cite{su2018splatnet}     &                                       &                                         &                                                             &                                                           &   22.8                               & 66.6                              & 0.0                                   & 0.0                                      & 0.0                                 & 0.0                                         & 0.0                                  & 0.0                                     & 0.0                                        & 70.4                               & 0.8                                   & 41.5                                   & 0.0                                        & 68.7                                   & 27.8                                & 72.3                                     & 35.9                                & 35.8                                  & 13.8                               & 0.0                                          \\
TangentConv \cite{Tat2018}  &                                          &                                         &                                                             &                                                           & 35.9                               & 86.8                              & 1.3                                   & 12.7                                     & 11.6                                & 10.2                                        & 17.1                                 & 20.2                                    & 0.5                                        & 82.9                               & 15.2                                  & 61.7                                   & 9.0                                        & 82.2                                   & 44.2                                & 75.5                                     & 42.5                               & 55.5                                  & 30.2                               & 22.2                                        \\
RandLA \cite{hu2019randla}       &                                          &                                         &                                                             &                                                           & 53.9                               & 94.2                              & 26.0                                  & 25.8                                     & \textbf{40.1}                                & 38.9                                        & 49.2                                 & 48.2                                    & 7.2                                        & 90.7                               & 60.3                                  & 73.7                                   & 20.4                                       & 86.9                                   & 56.3                                & 81.4                                     & 66.8                                & 49.2                                  & 47.7                               & 38.1                                        \\
KPConv \cite{thomas2019kpconv}       &                                          &                                         &                                                             &                                                           & \textbf{58.8}                               & \textbf{96.0}                              & 30.2                                  & \textbf{42.5}                                     & 33.4                                & \textbf{44.3}                                        & \textbf{61.5}                                 & \textbf{61.6}                                    & 11.8                                       & 88.8                               & 61.3                                  & 72.7                                   & 31.6                                       & \textbf{90.5}                                   & \textbf{64.2}                                & \textbf{84.8}                                     & \textbf{69.2}                                & 69.1                                  & \textbf{56.4}                               & 47.4                                        \\
\hline
SqueezeSeg \cite{8462926}   & \multirow{2}{*}{ 64 $\times$ 2048 px }   & \textbf{84.7}                           & -                                                           & -                                                         & 29.5                               & 68.8                              & 16.0                                  & 4.1                                      & 3.3                                 & 3.6                                         & 12.9                                 & 13.1                                    & 0.9                                        & 85.4                               & 26.9                                  & 54.3                                   & 4.5                                        & 57.4                                   & 29.0                                & 60.0                                     & 24.3                                & 53.7                                  & 17.5                               & 24.5                                        \\
SqueezeSegV2 \cite{8793495} &                                          & 55.8                                    & \textbf{928.5} K                                            & \textbf{13.6} G                                           & 39.7                               & 81.8                              & 18.5                                  & 17.9                                     & 13.4                                & 14.0                                        & 20.1                                 & 25.1                                    & 3.9                                        & 88.6                               & 45.8                                  & 67.6                                   & 17.7                                       & 73.7                                   & 41.1                                & 71.8                                     & 35.8                                & 60.2                                  & 20.2                               & 36.3                                        \\ 
\hline
RangeNet21 \cite{milioto2019rangenet++}   & 64 $\times$ 2048 px                      & 21.7                                    & -                                                           & -                                                         & 47.4                               & 85.4                              & 26.2                                  & 26.5                                     & 18.6                                & 15.6                                        & 31.8                                 & 33.6                                    & 4.0                                        & 91.4                               & 57.0                                  & 74.0                                   & 26.4                                       & 81.9                                   & 52.3                                & 77.6                                     & 48.4                                & 63.6                                  & 36.0                               & 50.0                                        \\
% RangeNet53  \cite{milioto2019rangenet++}& 64 $\times$ 512 px                       & 45.5                                    & -                                                           & -                                                         & 39.3                               & 81.0                              & 9.9                                   & 11.7                                     & 19.3                                & 7.9                                         & 16.8                                 & 25.8                                    & 2.5                                        & 90.1                               & 49.9                                  & 69.4                                   & 2.0                                        & 76.0                                   & 45.5                                & 74.2                                     & 38.8                                & 62.7                                  & 25.5                               & 38.1                                        \\
RangeNet53++ & 64 $\times$ 512 px                       & 38.5                                    &                                                                  &                                                                & 41.9                               & 87.4                              & 9.9                                   & 12.4                                     & 19.6                                & 7.9                                         & 18.1                                 & 29.5                                    & 2.5                                        & 90.0                               & 50.7                                  & 70.0                                   & 2.0                                        & 80.2                                   & 48.9                                & 77.1                                     & 45.7                                & 64.1                                  & 37.1                               & 42.0                                        \\
% RangeNet53   & 64 $\times$ 1024 px                      & 25.0                                    & -                                                           & -                                                         & 45.4                               & 84.6                              & 20.0                                  & 25.3                                     & 24.8                                & 17.3                                        & 27.5                                 & 27.7                                    & 7.1                                        & 90.4                               & 51.8                                  & 72.1                                   & 22.8                                       & 80.4                                   & 50.0                                & 75.1                                     & 46.0                                & 62.7                                  & 33.4                               & 43.4                                        \\
RangeNet53++ & 64 $\times$ 1024 px                      & 23.3                                    & -                                                           & -                                                         & 48.0                               & 90.3                              & 20.6                                  & 27.1                                     & 25.2                                & 17.6                                        & 29.6                                 & 34.2                                    & 7.1                                        & 90.4                               & 52.3                                  & 72.7                                   & 22.8                                       & 83.9                                   & 53.3                                & 77.7                                     & 52.5                                & 63.7                                  & 43.8                               & 47.2                                        \\
RangeNet53   & 64 $\times$ 2048 px                      & 13.3                                    & 50.4 M                                                      & 377.1 G                                                   & 49.9                               & 86.4                              & 24.5                                  & 32.7                                     & 25.5                                & 22.6                                        & 36.2                                 & 33.6                                    & 4.7                                        & \textbf{91.8}                      & 64.8                                  & 74.6                                   & 27.9                                       & 84.1                                   & 55.0                                & 78.3                                     & 50.1                                & 64.0                                  & 38.9                               & 52.2                                        \\
RangeNet53++ & 64 $\times$ 2048 px                      & 12.8                                    & -                                                           & -                                                         & 52.2                               & 91.4                              & 25.7                                  & 34.4                                     & 25.7                                & 23.0                                        & 38.3                                 & 38.8                                    & 4.8                                        & 91.8                               & \textbf{65.0}                                  & \textbf{75.2}                                   & 27.8                                       & 87.4                                   & 58.6                                & 80.5                                     & 55.1                                & 64.6                                  & 47.9                               & 55.9                                        \\ 
\hline
PolarNet \cite{zhang2020polarnet}     & 480 $\times$ 360 $\times$ 32             & 6.7                                     & 13.6 M                                                      & 135.0 G                                                   & 54.3                               & 93.8                              & \textbf{40.3}                                  & 30.1                                     & 22.9                                & 28.5                                        & 43.2                                 & 40.2                                    & 5.6                                        & 90.8                               & 61.7                                  & 74.4                                   & 21.7                                       & 90.0                                   & 61.3                                & 84.0                                     & 65.5                                & 67.8                                  & 51.8                               & 57.5                                        \\ 
\hline
MPF (ours)   & 64 $\times$ 512 px                       & 33.7                                    & -                                                           & -                                                         & 48.9                               & 91.1                              & 22.0                                  & 19.7                                     & 18.8                                & 16.5                                        & 30.0                                 & 36.2                                    & 4.2                                        & 91.1                               & 61.9                                  & 74.1                                   & 29.4                                       & 86.7                                   & 56.2                                & 82.3                                     & 51.6                                & 68.9                                  & 38.6                               & 49.8                                        \\
MPF (ours)   & 64 $\times$ 1024 px                      & 28.5                                    & -                                                           & -                                                         & 53.6                               & 92.7                              & 28.2                                  & 30.5                                     & 26.9                                & 25.2                                        & 42.5                                 & 44.5                                    & 9.5                                        & 90.5                               & 64.7                                  & 74.3                                   & \textbf{32.0}                              & 88.3                                   & 59.0                                & 83.4                                     & 56.6                                & \textbf{69.8}                                  & 46.0                               & 54.9                                        \\
MPF (ours)   & 64 $\times$ 2048 px                      & \underline{20.6}                                    & \underline{3.18 M}                                                      & \underline{27.0 G}                                                    & 55.5                               & 93.4                              & 30.2                                  & 38.3                                     & 26.1                                & 28.5                                        & 48.1                                 & 46.1                                    & \textbf{18.1}                                       & 90.6                               & 62.3                                  & 74.5                                   & 30.6                                       & 88.5                                   & 59.7                                & 83.5                                     & 59.7                                & 69.2                                  & 49.7                               & \textbf{58.1}                                        \\
\hline
\end{tabular}

\end{adjustbox}
\end{center}
\caption{mIoU scores on SemanticKITTI test set \protect\footnotemark. Our proposed MPF utilizes smaller number of parameters compared to projection-based methods Rangenet53++ \protect\cite{milioto2019rangenet++} and PolarNet \protect\cite{zhang2020polarnet} while maintaining higher segmentation results.}
\label{tab:results}
\end{table*}

%pseudocode
\begin{algorithm}[ht]
\footnotesize
\SetAlgoLined
\textbf{Post-Processing}\\

\textbf{Parameters} \\
    Number of classes: $C$ \\
    Number of points: $N$ \\
    Size of the projection image: $H$x$W$ \\
    Size of the sliding window : $K$x$K$ \\
    Standard deviation for the gaussian function: $\sigma$ \\
    \texttt{\\}

\textbf{Data} \\
    Point cloud coordinates $P_{xyz}$. Size = $N$x$3$ \\
    Projection image $I_{xyz}$. Size = $H$x$W$x$3$ \\
    Output of the segmentation network $I_{softmax}$. Size = $H$x$W$x$C$ \\
    \texttt{\\}
    
\textbf{Output} \\
    $Scores$. Class scores for each point in the original cloud. Size = $N$x$C$ \\
\texttt{\\}
    
\textbf{Algorithm} \\

    \ForEach {$i \in [1:N]$}
    {
        // Get the pixel to which this point is projected \\
        $u, v = get\_projection\_indices(P_{xyz}[i])$ \\

        // Initialize all class scores for the i'th points to zeros \\
        $Scores[i] = zeros(C)$ \\
        // Initialize number of non-sparse pixels to zero  \\
        $M = 0$ \\

        // Loop over pixels currently inside the sliding window  \\
        \ForEach {$u' \in [u-\floor{k/2}:u+\floor{k/2}]$}
        {
            \ForEach {$v' \in [v-\floor{k/2}:v+\floor{k/2}]$}
            {
                \If {$I_{xyz}[u',v']$ is not sparse}
                {
                    $d = get\_distance(P_{xyz}[i], I_{xyz}[u', v'])$ \\
                    $weight = exp(- d^2 / 2\sigma^2)$ \\
                    $Scores[i] += weight * I_{softmax}[u', v']$ \\
                    $M += 1$  \\
                }
            }
        }
        
        $Scores[i] = Scores[i] / M$ \\
    }
    
    return $Scores$

\caption{Post-Processing Algorithm}
\label{post_processing_algorithm}
\end{algorithm}

\subsection{Post-Processing}    \label{postprocessing}
The goal of the Post-Processing step is to get semantic labels for all points in the input 3D point cloud based on the semantically-segmented images produced by the Spherical and Bird's-Eye View Models. The segmentation results for each pixel are softmax probability scores for each of the 20 possible classes. The segmented 3D point cloud is computed as follows: each 3D point is projected onto the 2D segmented image and a 2D square window centered around the projection location is calculated. Then, a weighted vote over all classes is performed by computing a weighted sum of the softmax probabilities of all pixels inside the window, where the weights are inversely proportional to the distance between the 3D point under consideration and the 3D points represented by pixels in the window. In particular, we use a Gaussian function with zero mean and fixed standard deviation to compute the weight corresponding to each distance. The output of this step is a vector of scores for each point in the 3D point cloud. Finally, the score vector is normalized by dividing by the number of points that contributed in the voting. This step is necessary because both views have sparse pixels and the number of pixels inside a window can vary considerably from one view to another. The details of the algorithm are shown in Algorithm \ref{post_processing_algorithm}. During implementation, we eliminated the use of all loops and used fully-vectorized code which run on GPU for fast processing. Our proposed post-processing is similar to KNN post-processing in \cite{milioto2019rangenet++} however it uses soft voting with softmax probabilities instead of hard voting and takes the vote of non-sparse pixels only.

\subsection{Fusion}
After post-processing the outputs of the spherical and bird's-eye networks, we get two vectors of scores for each point, one vector for each view. These vector are simply added  to get the final score vector for each 3D point. The class that has highest score is selected as the predicted label.

\section{Experimental Evaluation}

\subsection{Datasets}
We trained both Spherical View and Bird's-Eye View networks on the SemanticKITTI dataset \cite{behley2019iccv} which provided point-wise semantic label annotations for all scans in the KITTI odometry dataset \cite{geiger2012cvpr}. The dataset consists of over 43,000 $360^{\circ}$ LiDAR scans, divided into 11 training sequences for which ground-truth annotations are provided and 11 test sequences. We used sequence $08$ as our validation set and trained our networks on the other 10 sequences.

\subsection{Training Configuration}

\subsubsection{Spherical View Model}

The Spherical View Model was trained from scratch using a combined objective function of Focal \cite{lin2017focal} and Lovász-Softmax \cite{berman2018lovasz} losses:
\[L_{spherical\ view\ model} = L_{focal} + L_{lovasz} \]
\[L_{focal} = -\sum_n\sum_i{(1 - p_{n,i})^\gamma log(p_{n,i})} \]
where $p_{n,i}$ is the probability of the ground-truth class at image $n$ and pixel $i$ and $\gamma$ is the focusing factor. For optimization, SGD with 0.9 momentum, 0.0001 weight decay and mini-batch of 8 was used. We also used Cosine Annealing scheduler with warm restart \cite{loshchilov2016sgdr} for 5 cycles with learning rate that starts at 0.05 and decreases till 0, and cycle length of 30 epochs. Following similar works, the model was trained with image sizes of 64x512, 64x 1024 and 64x2048.
\footnotetext{We included only peer-reviewed works in the study however there are other interesting approaches that can be easily incorporated in our framework such as SalsaNext \cite{cortinhal2020salsanext} and SqueezeSegV3 \cite{xu2020squeezesegv3}.}
\subsubsection{Bird's-Eye View Model}

To train the Bird's-Eye View Model, we used SGD with one-cycle \cite{DBLP:journals/corr/abs-1708-07120} learning and momentum annealing strategy. Learning rate was cycled between $0.001$ and $0.1$, while momentum was cycled inversely to learning rate between $0.85$ and $0.95$. The model was trained for 30 epochs using cross entropy loss and Lovász-Softmax loss:
\[L_{bird's-eye\ view\ model} = L_{cross\ entropy} + L_{lovasz} \]
\[L_{cross\ entropy} = -\sum_n\sum_i{log(p_{n,i})} \]
where $p_{n,i}$ is the probability of the ground-truth class at image $n$ and pixel $i$. The model was trained only using images of size 256x256.

\begin{figure*}[t]
    \centering
    \begin{tabular}{ccc}
    \includegraphics[width = 13em]{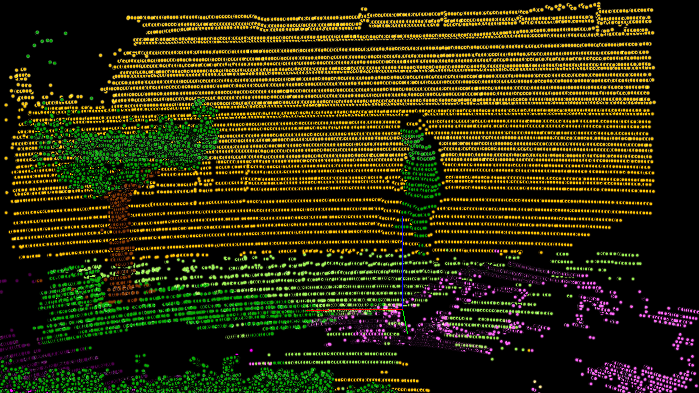} &
    \includegraphics[width = 13em]{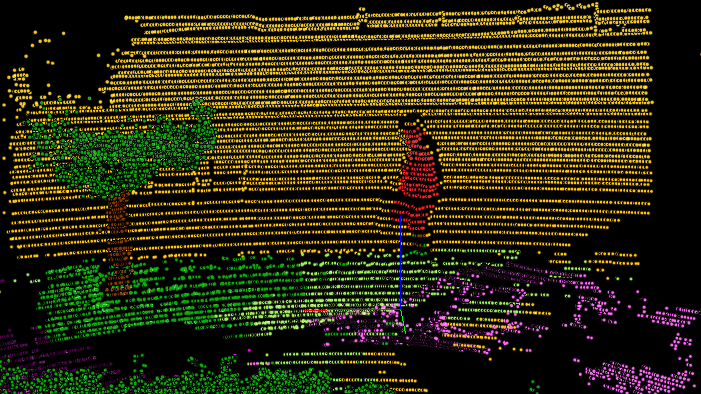} &
    \includegraphics[width = 13em]{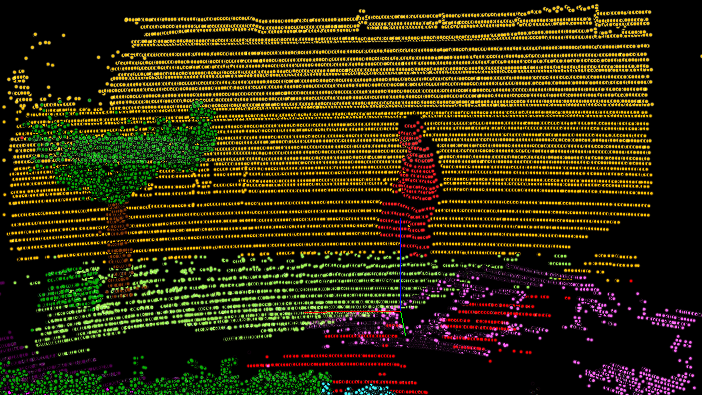}\\
    \includegraphics[width = 13em]{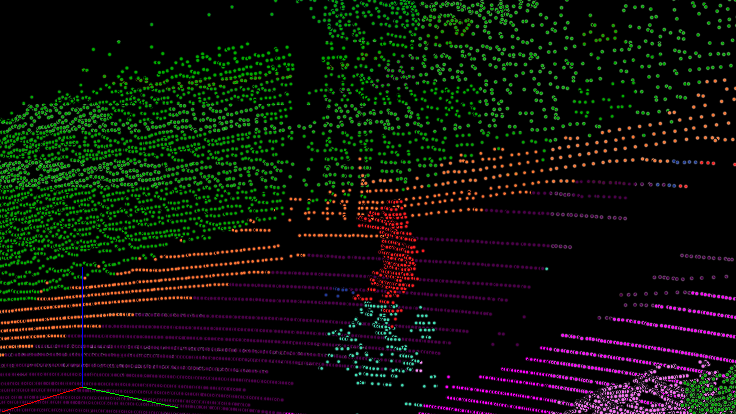} &
    \includegraphics[width = 13em]{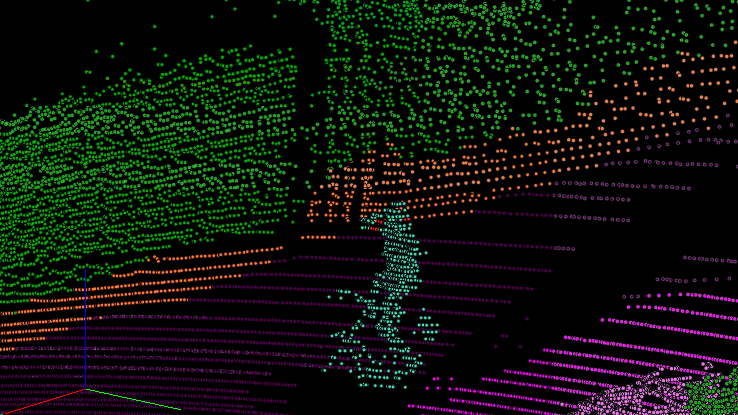} &
    \includegraphics[width = 13em]{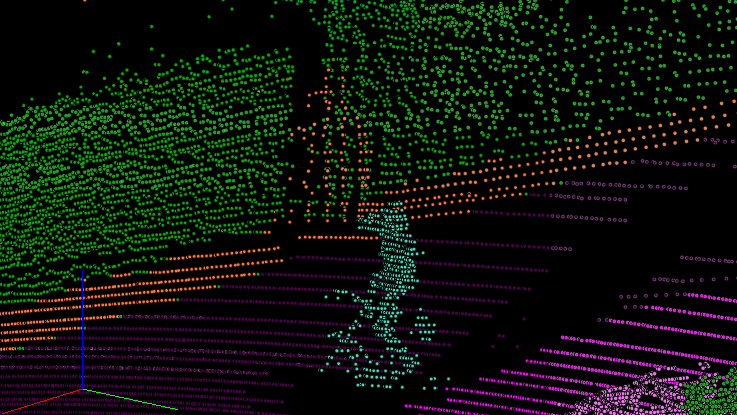}\\
    \subfloat[RangeNet53++]{\includegraphics[width = 13em]{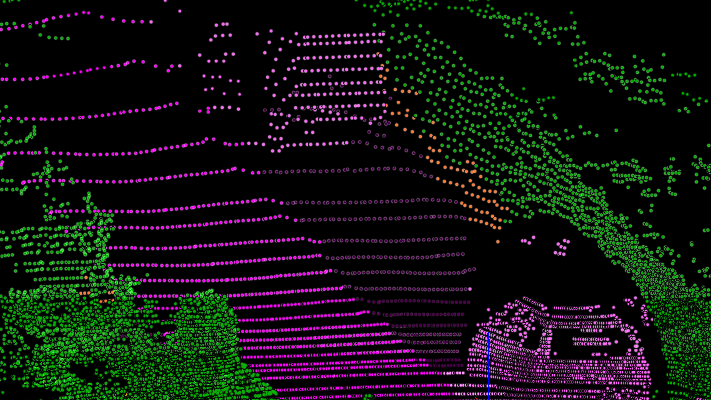}} &
    \subfloat[MPF (ours)]{\includegraphics[width = 13em]{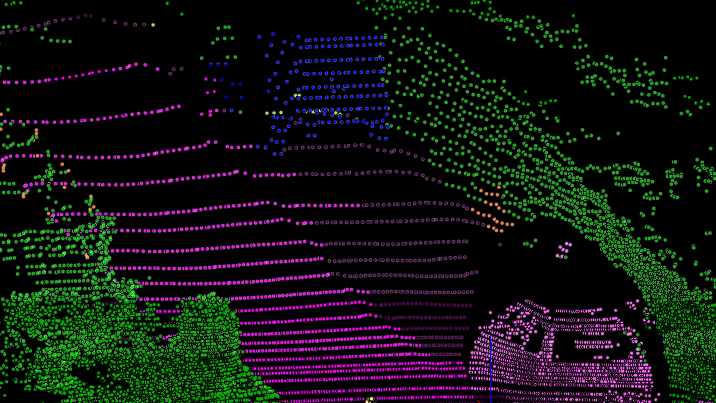}} &
    \subfloat[Ground Turth]{\includegraphics[width = 13em]{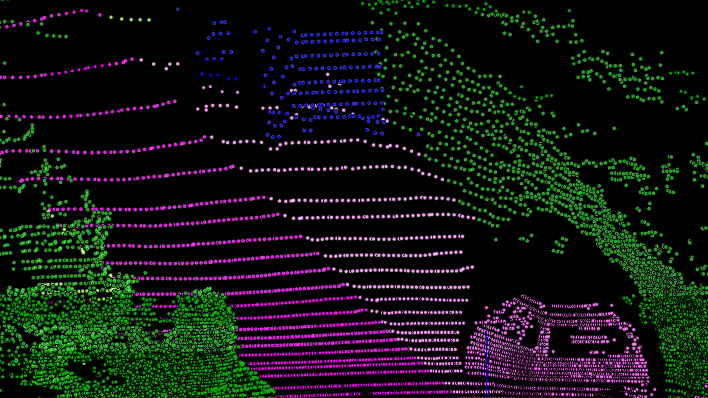}}
    \end{tabular}
    \caption{Qualitative segmentation results on SemanticKITTI validation sequence 8 that compare our MPF framework against RangeNet53++. Top: Our framework correctly segmented the 'person' labeled in red. Middle: Our framework segmented the 'bicyclist' object, a class rarely representated in the dataset, much better than RangeNet53++. Bottom: the MPF framework correctly segmented the 'other-vehicle' object (in upper middle location), which was completely missed by RangeNet53++.}
    \label{fig:my_label}
\end{figure*}

\subsubsection{Post-Processing}
For post-processing, we used square kernels of size 3 for both views to extract local windows centered at projected pixels. Network predictions for sparse pixels (pixels that have no corresponding 3D points) are ignored. Then to compute the weight for each pixel, a Gaussian function is used as described in Section \ref{postprocessing} with $\sigma = 1$ to compute the weights corresponding to different distances.

\subsection{Data Augmentation}
\label{data_augment}
Data augmentation is considered an effective tool to improve model generalization to unseen data. We therefore apply Spherical-View augmentation and Bird’s-Eye View augmentation. In the former, the 3D point cloud is processed by pipeline of four randomly executed (with 0.5 probability) Affine transformations: translation parallel to y-axis, rotation about z-axis, scaling around the origin and flipping around y-axis. Augmentation is also applied after projecting the 3D point cloud onto 2D space. A CoarseDropout function by \cite{imgaug} is used to drop pixels randomly by 0.005 probability and the image is cropped to half on the horizontal axis starting from a randomly sampled coordinate. 

In Bird’s-Eye View augmentation and prior to projecting the point cloud on the $x-y$ plane, the cloud is transformed by applying random rotation around the $z$ axis, scaling by a uniformly sampled factor, translating in the $x$ and $y$ directions and finally a random noise sampled from a normal distribution with $0$ mean and $0.2$ standard deviation is applied to the $z$ channel. Each of these operations is applied with probability of $0.5$.

\subsection{Results}
This section describes the performance of the proposed Multi-Projection Fusion (MPF) framework. Table \ref{tab:results} presents the quantitative results obtained by the proposed approach versus state-of-the-art point cloud semantic segmentation methods on the  SemanticKITTI test set over 19 classes. SemanticKITTI uses Intersection-over-Union (IoU) metric to report per-class score:
\begin{equation}
    IoU = \frac{|P \cap G|}{|P \cup G|}
\end{equation}
where $P$ and $G$ are class points predictions and ground truth, respectively. The mean IoU (mIoU) over all classes is also reported. The scans per second rate are reported by measuring combined projection and inference time (unlike PolarNet \cite{zhang2020polarnet} which reports inference time only) on a single NVIDIA GeForce GTX 1080 Ti GPU card. The results demonstrate that the proposed MPF framework achieves the highest mIoU score across all baseline projection-based methods while also having higher scans per second rate and less parameters compared to RangeNet++ \cite{milioto2019rangenet++} and PolarNet \cite{zhang2020polarnet}. Although 3D-methods achieves the highest mIOU scores, it lags significantly in running time as shown in \cite{tang2020searching} and \cite{hu2019randla}.

Figure \ref{fig:qualitative_result} show qualitative examples from the SemanticKITTI validation sequence where our proposed approach outperforms RangeNet53++ \cite{milioto2019rangenet++} in segmenting objects located far from LiDAR position. This is attributed to using two independent complementary projections and intelligently fusing segmentation results of each projection. It is worth mentioning that data augmentation described in Section \ref{data_augment} improved validation mIoU by $2.7\%$ for spherical view model, $7.7\%$ for bird's-eye view model and an overall improvement by $7.2\%$ as shown in Table \ref{tab:ablation_augmentation}.

\subsection{Ablation Study}
\subsubsection{Post-Processing}
\begin{figure}[t]
    \centering
    \includegraphics[width=\linewidth]{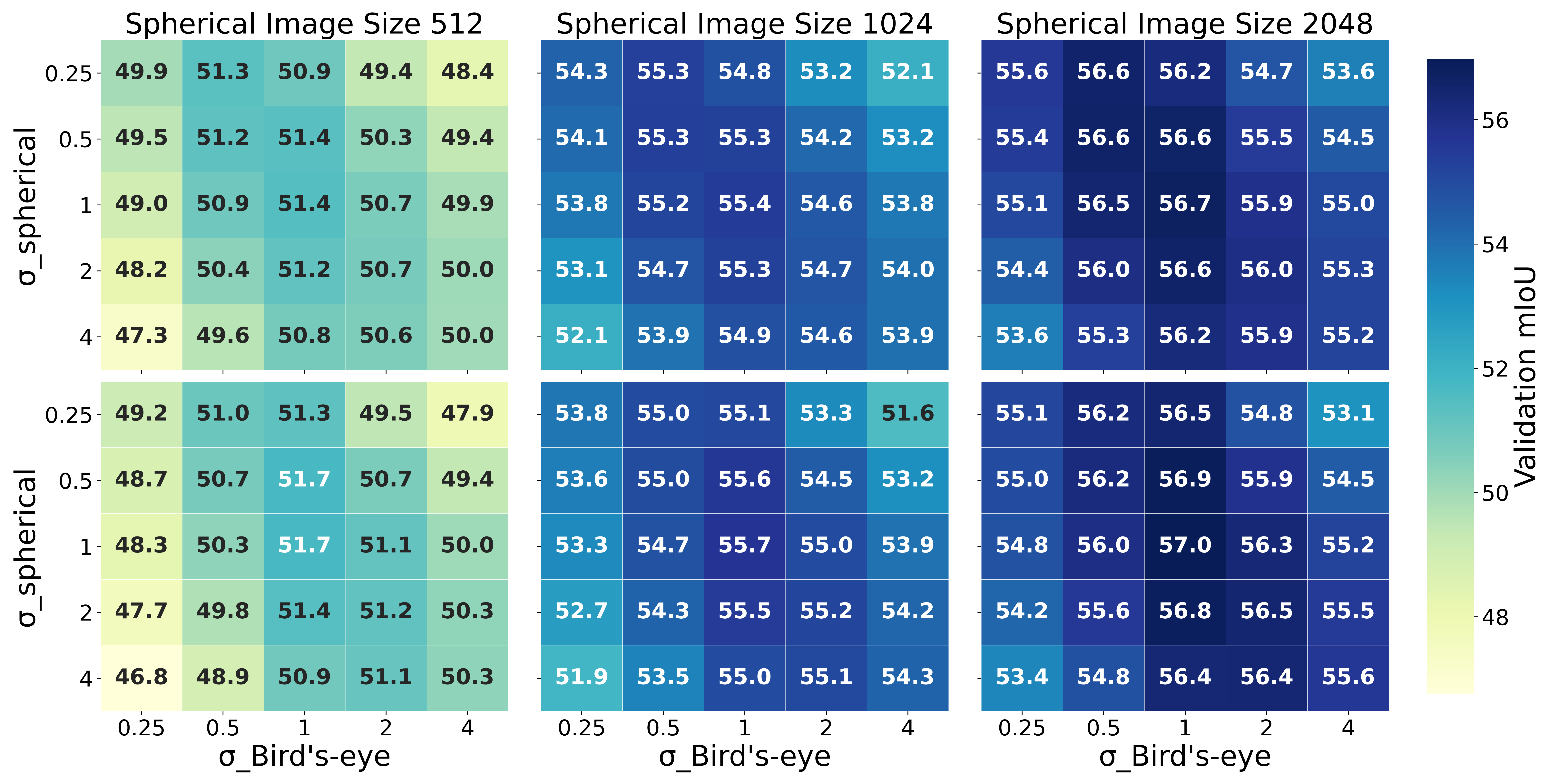}
    \caption{Post-processing $\sigma$ values of spherical and bird's-eye views against validation mIoU. Top: Euclidean distance. Bottom: Manhattan distance.}
    \label{fig:postprocessing_ablation}
    
\end{figure}

We studied the effect of different standard deviation values on the performance of the Gaussian function used presented in Algorithm \ref{post_processing_algorithm}. A grid search was used to jointly compute the best values for the standard deviation used in both spherical view and bird's-eye view post-processing. As shown in Figure \ref{fig:postprocessing_ablation}, the results show that setting all standard deviation values to $1$ when using Manhattan distance consistently yields the best results. We also tried larger sliding window sizes but it did not improve the score and reduced the overall FPS of our framework. The best configurations from this study were used for test submission.

\begin{table}
\centering
\begin{tabular}{l|l|l} 
\hline
Model            & \begin{tabular}[c]{@{}l@{}}Without \\Augmentation\end{tabular} & \begin{tabular}[c]{@{}l@{}}With~\\Augmentation\end{tabular}  \\ 
\hline
Spherical view~  & 39.9                                                           & \textbf{42.6}                                                \\ 
\hline
Bird's-eye view~ & 33.6~                                                          & \textbf{41.3}                                                \\
\hline
MPF~ & 44.5~                                                          & \textbf{51.7}                                                \\
\hline
\end{tabular}
\caption{Ablation study on the effect of using augmentation on validation mIoU score. The image size used for spherical view model is 64x512.}
\label{tab:ablation_augmentation}
\end{table}

\subsubsection{Multi-Projection Fusion}

We conducted several ablation studies to demonstrate the efficacy of employing multiple projections as opposed to single projection. In the first experiment, we investigate the mIoU score of two established spherical projection models, SqueezeSeg \cite{8462926,8793495} and RangeNet \cite{noh2015learning}, with and without the incorporation of the bird's-eye projection in table \ref{tab:fusion_ablation}. The results demonstrate that fusion of information from more than one projection significantly enhances the obtained segmentation results despite using simple network model and projected low-resolution images for the bird's-eye view. It can thus be concluded that using multiple projections of the same point cloud does improve overall segmentation results by providing additional information for model adaptation.

\begin{figure}[t]
     \centering
     \begin{subfigure}[b]{0.5\linewidth}
         \centering
         \includegraphics[width=\linewidth]{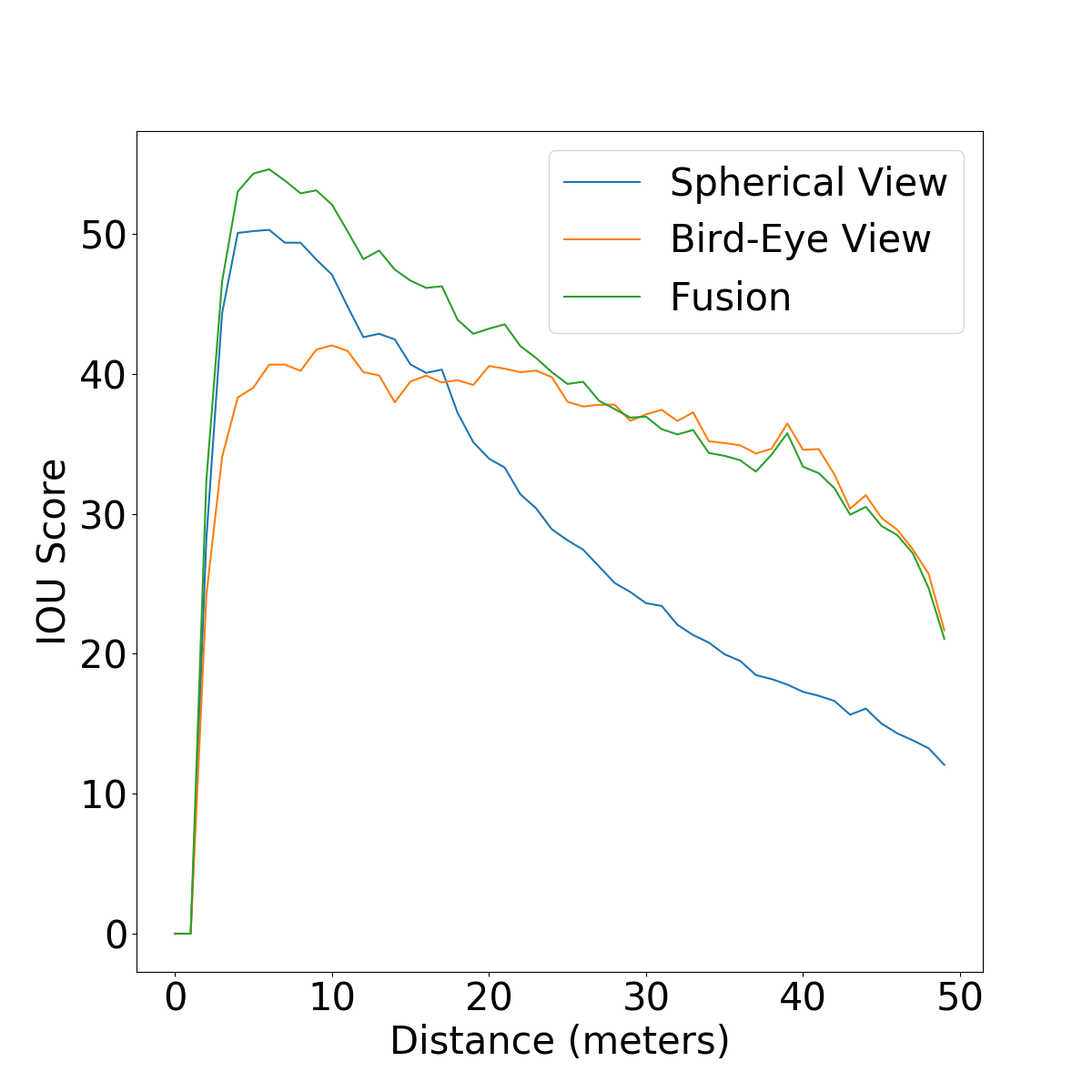}
        \caption{}
        \label{fig:fusion_ablation_1}
     \end{subfigure}%
     \begin{subfigure}[b]{0.5\linewidth}
         \centering
         \includegraphics[width=\linewidth]{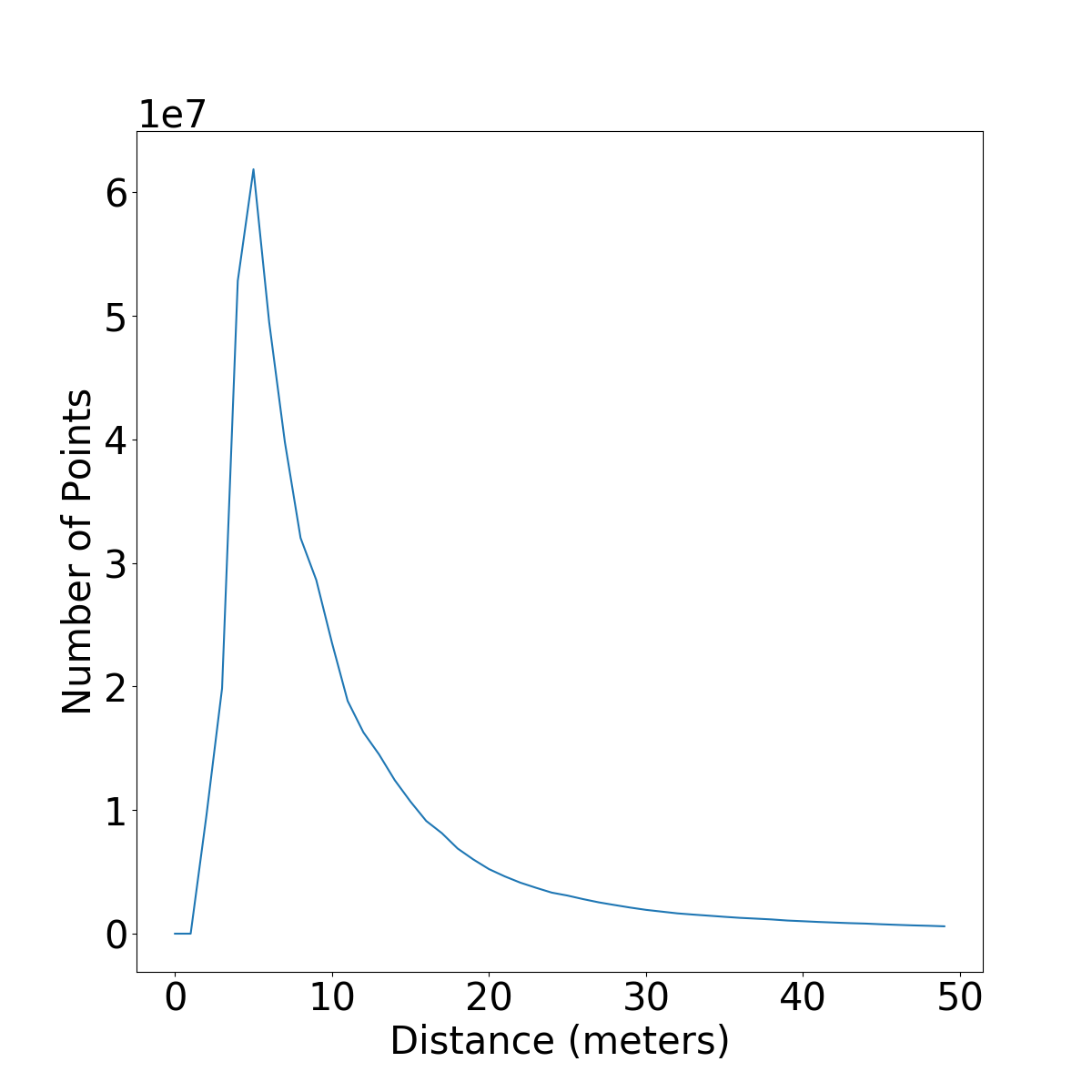}
        \caption{}
        \label{fig:fusion_ablation_2}
     \end{subfigure}
        \caption{(a) The mIoU score vs. distance for near and far points. Spherical view image of size 512 used in this experiment. (b) Number of points in the validation set vs. distance.}
        \label{fig:three graphs}
\end{figure}

The second experiment shows that fusion helps to improve mIoU score for both near and far points as seen in Figure \ref{fig:fusion_ablation_1}. Since the majority of LiDAR points are typically at distances $<$ 20 meters, as shown in Figure \ref{fig:fusion_ablation_2}, a slight improvement of framework performance for near distance points can have a significant impact on the overall IoU results. It also observed that the performance of point cloud segmentation using spherical view projections degrades for points farther away from the LiDAR position. Unlike spherical view images, the bird's-eye view images use Cartesian coordinates which means the pixels of the image correspond to uniform 3D elements whose spatial resolution does not change as we go farther from the LiDAR position which makes segmentation results independent of point distances.

% \begin{figure}[t]

% \begin{subfigure}
% \includegraphics[width=0.5\textwidth]{LightUnet_512 No Accumulation.png}

% \caption{The mIoU score versus distance. We used spherical view image size 512 for this experiment.}
% \label{fig:fusion_ablation_1}
% \end{subfigure}

% \begin{subfigure}
% \includegraphics[width=0.5\textwidth]{Num Points_512 No Accumulation.png}
% \caption{Number of points in the validation set versus distance.}
% \label{fig:fusion_ablation_2}
% \end{subfigure}

% \end{figure}

\begin{table}
\begin{center}
\begin{tabular}{l|c|c|c} 
\hline
\multicolumn{1}{c|}{Network} & Size & W/o fusion   & W fusion  \\ 
\hline
SqueezeSeg\cite{8462926}                & 2048 & 30.5  & \textbf{44.3}         \\
SqueezeSegV2\cite{8793495}              & 2048 & ~40.4 & \textbf{48.6}         \\ 
\hline
RangeNet53++\cite{milioto2019rangenet++}                & 512  & 37.5   & \textbf{44.4}         \\
RangeNet53++                   & 1024 & 36.5  & \textbf{42.2  }         \\
RangeNet53++                   & 2048 & 50.3  & \textbf{55.4}         \\ 
\hline
MPF (ours)                        & 512  & 42.6  & \textbf{51.7}         \\
MPF (ours)                        & 1024 & 48.5  & \textbf{55.7}         \\
MPF (ours)                        & 2048 & 50.7  & \textbf{57.0}         \\
\hline
\end{tabular}

\end{center}
\caption{Results of adding the bird's-eye projection to single spherical projection models on SemanticKITTI validation set.}
\label{tab:fusion_ablation}
\end{table}

\section{Conclusions and Future Work}

This paper presented a novel multi-projection fusion framework for point cloud semantic segmentation by using spherical and bird's-eye view projections and fusion of results using soft voting mechanism. The proposed framework achieves improved segmentation results over single projection methods while having higher throughput. Future work directions include combining both projections into a single multi-view unified model and investigating using more than two projections within the framework.

{\small
\bibliographystyle{ieee_fullname}
\bibliography{egpaper}
}

\end{document}